# Federated Learning with Multi-Partner OneFlorida+ Consortium Data for Predicting Major Postoperative Complications

Yuanfang Ren, PhD[a,b], Varun Sai Vemuri, BS[a], Zhenhong Hu, PhD[a,b], Benjamin Shickel, PhD[a,b], Ziyuan Guan, MS[a,b], Tyler J. Loftus, MD, PhD[a,c], Parisa Rashidi, PhD[a,d], Tezcan Ozrazgat-Baslanti, PhD[a,b*], Azra Bihorac, MD, MS[a,b*]

*These senior authors have contributed equally

[a] Intelligent Clinical Care Center, University of Florida, Gainesville, FL.

[b] Department of Medicine, Division of Nephrology, Hypertension, and Renal Transplantation, University of Florida, Gainesville, FL.

[c] Department of Surgery, University of Florida, Gainesville, FL.

[d] Department of Biomedical Engineering, University of Florida, Gainesville, FL.

**Corresponding to**: Azra Bihorac, MD, MS, Intelligent Clinical Care Center, Division of Nephrology, Hypertension, and Renal Transplantation, Department of Medicine, University of Florida, PO Box 100224, Gainesville, FL 32610-0224. Telephone: (352) 294-8580; Fax: (352) 392-5465; Email: abihorac@ufl.edu

Reprints will not be available from the authors.

**Conflicts of Interest and Source of Funding**: AB, PR, TOB, BS, and YR were supported R01 GM110240 from the National Institute of General Medical Sciences (NIH/NIGMS). This work was also supported in part by the National Center for Advancing Translational Sciences of the National Institutes of Health under the University of Florida Clinical and Translational Science Awards UL1 TR000064 and UL1 TR001427. The content is solely the responsibility of the authors and does not necessarily represent the official views of the National Institutes of Health. The funders had no role in design and conduct of the study; collection, management, analysis, and interpretation of the data; preparation, review, or approval of the manuscript; and decision to submit the manuscript for publication. The authors declare that they have no conflict of interests. AB and TOB had full access to all the data in the study and take responsibility for the integrity of the data and the accuracy of the data analysis.

**Abstract**

**Background**: This study aims to develop and validate federated learning models for predicting major postoperative complications and mortality using a large multicenter dataset from the OneFlorida Data Trust. We hypothesize that federated learning models will offer robust generalizability while preserving data privacy and security.

**Methods:** This retrospective, longitudinal, multicenter cohort study included 358,644 adult patients admitted to five healthcare institutions, who underwent 494,163 inpatient major surgical procedures from 2012-2023. We developed and internally and externally validated federated learning models to predict the postoperative risk of intensive care unit (ICU) admission, mechanical ventilation (MV) therapy, acute kidney injury (AKI), and in-hospital mortality. These models were compared with local models trained on data from a single center and central models trained on a pooled dataset from all centers. Performance was primarily evaluated using area under the receiver operating characteristics curve (AUROC) and the area under the precision-recall curve (AUPRC) values.

**Results:** Our federated learning models demonstrated strong predictive performance, with AUROC scores consistently comparable or superior performance in terms of AUROC and AUPRC across all outcomes and sites. Our federated learning models also demonstrated strong generalizability, with comparable or superior performance in terms of both AUROC and AUPRC compared to the best local learning model at each site.

**Conclusions:** By leveraging multicenter data, we developed robust, generalizable, and privacy-preserving predictive models for major postoperative complications and mortality. These findings support the feasibility of federated learning in clinical decision support systems.

## Introduction

In the United States, a staggering 40 to 50 million major surgeries are performed annually.[1] It is estimated that postoperative complications occur in up to 20% of cases on average[2], through rates can reach 75% for high risk procedures[3], contributing to increased morbidity, prolonged hospital length of stay, elevated healthcare costs, and even mortality.[4-6] While the 30-day postoperative mortality rate typically ranges from 1% to 4% for major surgeries, the 1-year mortality rate for high-risk populations can exceed 13%.[1,7] Accurate prediction of postoperative complication risk during the preoperative period is crucial for improving patient outcomes and optimizing resource allocation.

Assessing surgical risk requires the timely and precise aggregation of extensive clinical data. The integration of artificial intelligence with electronic health records (EHR) has advanced data digitization, thereby enhancing the use of machine learning tools for risk monitoring and diagnosis. Despite the promising developments in predicting these risks[8-10], current models are often limited to single hospitals, restricting their generalizability and reducing their effectiveness in diverse clinical settings. Multicenter data offers a solution to this limitation by enabling the development of more robust and generalizable predictive models. The rich diversity of patient populations and clinical practices captured in multicenter datasets can significantly enhance the performance and applicability of predictive models. However, the use of such data raises significant privacy concerns, as patient information must be securely managed and protected.

Federated learning addresses these concerns by allowing multiple institutions to collaboratively train machine learning models without sharing raw patient data. Instead, model updates are aggregated centrally, ensuring data privacy while leveraging the rich and diverse data available across institutions. Although federated learning has been successfully applied to predict outcomes such as acute kidney injury (AKI) stage, adverse drug reactions, hospitalizations, and COVID-19 mortality, its application for predicting surgical postoperative outcomes has not been thoroughly explored. [11-15] For example, Ren at al.[16] developed federated learning models for eight major postoperative complications and in-hospital mortality using data from two University of Florida Health (UFH) hospitals. However, the limited number of hospitals from the same system restricts the generalizability of these findings.

To address this gap, our study aims to develop and validate federated learning models for predicting major postoperative complications and mortality using a large multicenter dataset from the OneFlorida Data Trust—a clinical data research network comprising 14 health systems. Our hypothesis was that our federated learning model trained on distributed data would have better generalizability while simultaneously preserving data privacy and security.

## Methods

### *Study Design and Participants*

We obtained longitudinal EHR data for 1,455,294 hospitalized patients admitted to healthcare institutions within the OneFlorida+ network between January 1, 2012 and April 30, 2023, comprising 81,421,419 admissions (including historical admissions). We excluded outpatient admissions, patients under 18 years of age at the time of admission, patients who did not undergo major surgery, and those with end-stage renal disease (Figure 1). Our final cohort included 494,163 encounters from 358,644 patients from 5 partners (Partners 1, 2, 3, 4, and 6). The dataset includes demographics, vital signs, laboratory values, medications, diagnosis, and procedure codes for all admissions. The study was approved by the University of Florida Institutional Review Board and Privacy Office (IRB# IRB202300641) as an exempt study with a waiver of informed consent.

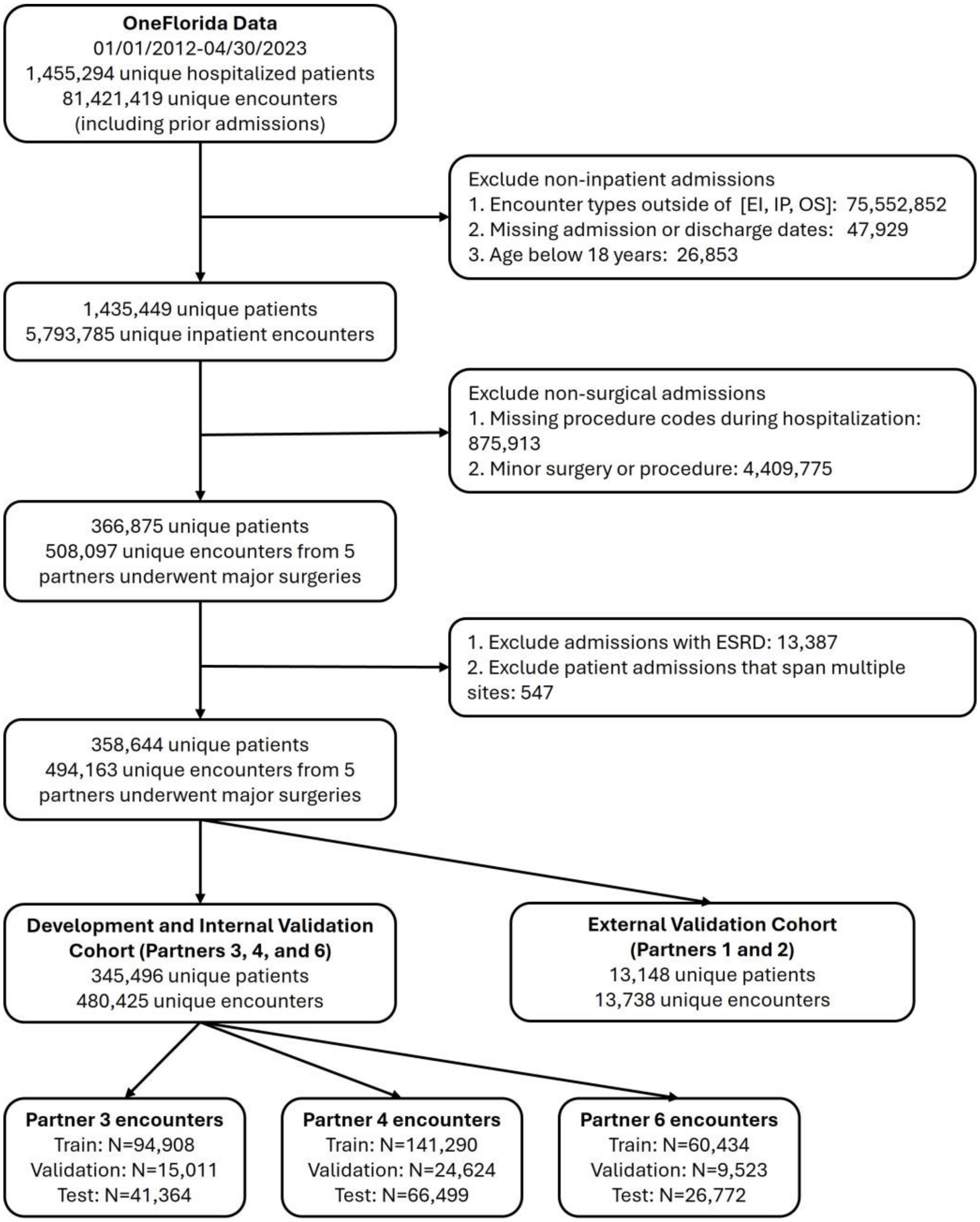

**Figure 1. Cohort selection and exclusion criteria**

We developed predictive models using three distinct learning paradigms: local learning, central learning and federated learning. Local learning develops models using data from a single center, which can lead to models with limited generalizability. Central learning, which pools data from multiple centers to develop models, allows for more generalized models but raises potential privacy concerns. Federated learning addresses these challenges by training the model locally at each center and aggregating the computed models on a central server without exchanging actual data. Considering the diversity of patient populations and the volume of available data, we selected three sites (Partners 3, 4, and 6) for model development and internal validation. The remaining two sites (Partners 1 and 2) were reserved for external validation. Following the Transparent Reporting of a multivariable prediction model for Individual Prognosis or Diagnosis (TRIPOD) guidelines under the type 2b analysis category, we chronologically split each dataset at the three development sites by admission dates into three cohorts: training (60% of observations), validation (10% of observations) and the test (30% of observations) cohorts (Figure 1). This approach mitigates potential adverse effects of dataset drift due to changes in clinical practice or patient populations over time. We ensured that encounters from the same patient stayed within the same cohort and the same site. We developed the model using the training cohort, tuned hyper-parameters and selected models using validation cohort, and internally validated the model using the test cohort.

*Identification of Major Surgery and Outcomes*

We identified major surgery using Current Procedural Terminology (CPT) codes and associated relative value units (RVUs) following the methodology in study[17]. A CPT code represents major surgery if the linked RVU includes an intraoperative portion and has an indicator of major surgery (Ren at al.[17] for more details). When a patient underwent multiple surgeries during one admission, only the surgery with the highest intraoperative working units was included in the analysis. Due to the unavailability of detailed surgical information, we used the dates associated with the CPT codes to mark the start and end dates of the surgery, as precise surgery start and end times were not accessible.

The primary outcomes were three postoperative complications: postoperative intensive care unit (ICU) admission, postoperative mechanical ventilation (MV), AKI and in-hospital mortality. ICU admissions and MV were identified using diagnosis and procedure codes. We used our previously developed and validated computable phenotype algorithms[18] to determine the presence of AKI. In-hospital mortality was determined using the date of death. These four outcomes were tracked from the beginning of surgery (including the surgery date) until discharge. For more details, we refer readers to Ren at al.[17]

*Predictors and Data Preprocessing*

Following our previous study[17], we used 99 routinely collected preoperative features, along with historical admission information. These features included demographics and socioeconomic factors (e.g., insurance type), admission characteristics (e.g., admission source), operative information (e.g., procedure code, provider specificity), comorbidities, historical medications and preoperative laboratory measurements. For the complete list of features, see Ren et al.[17]

We preprocessed the feature variables by removing outliers, imputing missing values, and standardizing the data.[9,16,19] We removed unreasonable values as determined by medical experts, as well as values in the top and bottom 1% of the distribution. We addressed missing values by creating a 'missing' category for missing categorical variables and imputing the missing continuous variables with the median value from the training cohort. We standardized the continuous variables using min-max normalization techniques. For federated learning, all partners employed the same standardization scaler, but other preprocessing steps were independently carried out at each site.

*Model Architecture*

Following the model architecture described in previous studies[9,16], we developed a deep learning model to predict the risk of postoperative complications. Features in this model were divided into continuous, binary, and high-cardinality types, and processed through neural networks suited to each type. Continuous and binary features were processed using fully connected layers, while high-cardinality features were handled with embedding layers followed by fully connected layers. The latent representations from the three neural networks were merged through a fully connected layer. This combined output was then directed into four separate branches, each corresponding to one of the four outcomes, to estimate the risk probabilities. We refer readers to Shickel et al.[9] for architecture details.

We trained the model using local learning, central learning and federated learning respectively. For federated learning, we employed three different algorithms: FedAvg[20], FedProx[21] and SCAFFOLD[22]. FedAvg averages model weights from multiple clients, while FedProx extends FedAvg by penalizing a client’s local weights based on the deviation from the global model. SCAFFOLD specifically addresses the problems of non-Independently and Identically Distributed (Non-IID) data in federated learning environments. In addition to these federated learning algorithms tailored to neural network models, we also developed an XGBoost (eXtreme Gradient Boosting) model, implemented using the federated learning algorithm Federated XGBoost[23].

*Statistical Analysis*

We conducted a sensitivity analysis to explore the combination of federated learning and local learning by fine-tuning the federated learning model at each site using a personalized feature: the surgeon's identity. We froze the weights of the layers before the last output layer, added an additional embedding layer for the surgeon's identity feature, and concatenated the feature representation from the federated learning model with the personalized features to generate new predictions.

We evaluated model performance primarily using the area under the receiver operating characteristic curve (AUROC) and the area under the precision-recall curve (AUPRC). For the sensitivity analysis, we also assessed performance using sensitivity, specificity, positive predictive value (PPV), and negative predictive value (NPV). We conducted 1000 bootstrap samplings to obtain 95% confidence intervals (CIs) for all performance metrics. For comparison of clinical characteristics and outcomes, we used the $\chi 2$ test for categorical variables and the Mann-Whitney U test for continuous variables. The threshold for statistical significance was less than 0.05 for 2-sided tests, and P values for the family-wise error rate resulting from multiple comparisons were adjusted using the Bonferroni correction. Analysis was conducted using Python version 3.10, NVFlare version 2.7 and R version 4.5.2.

## Results

### *Patient Clinical Characteristics and Outcomes*

Across all training cohorts for model development, significant differences were observed in both demographic characteristics and outcome distributions (Table 1). In the training cohort of 73,458 adult patients who underwent 94,908 major surgeries from Partner 3, the mean age (standard deviation, SD) was 56 (20) year; 37,455 (51%) were female; 53,158 (72%) were White and 15,330 (21%) African American; and 3,540 (5%) were Hispanic. The prevalence of outcomes was 15% for postoperative ICU admission, 6% for postoperative MV, 10% for AKI, and 2% for in-hospital mortality. The Partner 4 training cohort included 85,623 patients who underwent 141,290 major surgical procedures. Compared to Partner 3, this cohort had a higher mean (SD) age at 61 (19) years, a higher percentage of female patients (53%), a higher percentage of white patients (77%), a significantly higher percentage of Hispanic patients (44%), and significantly lower prevalence of outcomes, including 2% for ICU admission, 1% for postoperative MV, 1% for AKI and 0.1% for in-hospital mortality. The Partner 6 training cohort included 48,542 patients who underwent 60,434 major surgical procedures. Compared to the other two partners, this cohort had an intermediate mean (SD) age at 57 (18) years, a higher percentage of African-American patients (24% vs 21% in Partner 3 and 14% in Partner 4), a significantly lower percentage of Hispanic patients (1%), and relatively lower prevalence of most outcomes, including 6% for ICU admission, 2% for MV, 15% for AKI, and 1% for in-hospital mortality.

The external validation cohort included 13,148 patients who underwent 13,738 major surgical procedures. Compared to the model development cohorts, this cohort had younger patients (mean [SD] age, 50 [20] years), a higher percentage of female patients (67%), similar percentage of White and African-American patients to Partner 3 and Partner 6, and significantly lower prevalence of outcomes, including 0.3% for ICU admission, 0.3% for MV, 2% for AKI, and 0.1% for in-hospital mortality.  For each partner, its internal validation cohort also exhibited significant differences in both demographic characteristics and outcome distributions compared to its training cohort for model development.

**Table 1. Patient baseline characteristics**

| **Variable** | **Partner 3** | | **Partner 4** | | **Partner 6** | | **External** | **P[b]** |
|---|---|---|---|---|---|---|---|---|
| | **Train** | **Test** | **Train** | **Test** | **Train** | **Test** | **Test** | |
| Number of patients, n | 73,458 | 36,442 | 85,623 | 42,818 | 48,542 | 24,001 | 13,148 | |
| Number of surgeries, n | 94,908 | 41,364 | 141,290 | 66,499 | 60,434 | 26,772 | 13,738 | |
| Age in years, mean (SD)[c] | 56 (20) | 57 (18)[a] | 61 (19) | 59 (17)[a] | 57 (18) | 57 (17)[a] | 50 (20) | <0.001 |
| Sex, n (%)[c] | | | | | | | | |
| Male | 36,003 (49) | 18,953 (52)[a] | 40,373 (47) | 20,802 (49)[a] | 21,994 (45) | 11,709 (49)[a] | 4,296 (33) | <0.001 |
| Female | 37,455 (51) | 17,489 (48)[a] | 45,250 (53) | 22,016 (51)[a] | 26,548 (55) | 12,292 (51)[a] | 8,852 (67) | <0.001 |
| Race, n (%)[c,d,e] | | | | | | | | |
| White | 53,158 (72) | 26,512 (73) | 66,015 (77) | 34,531 (81)[a] | 34,604 (71) | 16,704 (70)[a] | 9,471 (72) | <0.001 |
| African American | 15,330 (21) | 7,166 (20)[a] | 11,876 (14) | 5,531 (13)[a] | 11,637 (24) | 5,722 (24) | 2,944 (22) | <0.001 |
| Other | 4,183 (6) | 2,369 (7)[a] | 1,536 (2) | 843 (2)[a] | 2,099 (4) | 1,465 (6)[a] | 254 (2) | <0.001 |
| Missing | 787 (1) | 395 (1) | 6,196 (7) | 1,913 (4)[a] | 202 (0.4) | 110 (0.5) | 479 (4) | <0.001 |
| Ethnicity, n (%)[c,d] | | | | | | | | |
| Non-Hispanic | 69,113 (94) | 33,568 (92)[a] | 45,386 (53) | 20,949 (49)[a] | 45,045 (93) | 21,919 (91)[a] | 11,171 (85) | <0.001 |
| Hispanic | 3,540 (5) | 2,159 (6)[a] | 37,843 (44) | 19,779 (46)[a] | 684 (1) | 485 (2)[a] | 948 (7) | <0.001 |
| Missing | 805 (1) | 715 (2)[a] | 2,394 (3) | 2,090 (5)[a] | 2,813 (6) | 1,597 (7)[a] | 1,029 (8) | <0.001 |
| Insurance [c] | | | | | | | | |
| Medicare | 28,822 (39) | 15,677 (43)[a] | 23,967 (28) | 8,707 (20)[a] | 15,178 (31) | 5,378 (22)[a] | 1,534 (12) | <0.001 |
| Private | 21,994 (30) | 11,191 (31)[a] | 49,953 (58) | 27,780 (65)[a] | 18,797 (39) | 9,074 (38)[a] | 10,167 (77) | <0.001 |
| Medicaid | 15,231 (21) | 6,520 (18)[a] | 1,724 (2) | 552 (1)[a] | 1,761 (4) | 2,147 (9)[a] | 897 (7) | <0.001 |
| Uninsured | 7,411 (10) | 3,054 (8)[a] | 9,979 (12) | 5,779 (14)[a] | 12,806 (26) | 7,402 (31)[a] | 550 (4) | <0.001 |
| Complications, n (%)[f] | | | | | | | | |
| ICU admission | 13991 (15) | 10469 (25)[a] | 2325 (2) | 2608 (4)[a] | 3849 (6) | 1975 (7)[a] | 41 (0.3) | <0.001 |
| MV | 6104 (6) | 4809 (12)[a] | 1577 (1) | 1584 (2)[a] | 1124 (2) | 1212 (5)[a] | 40 (0.3) | <0.001 |
| AKI | 9618 (10) | 5802 (14)[a] | 1364 (1) | 1207 (2)[a] | 8959 (15) | 4810 (18)[a] | 287 (2) | <0.001 |
| In-hospital mortality | 1976 (2) | 866 (2) | 130 (0.1) | 62 (0.1) | 706 (1) | 390 (1)[a] | 11 (0.1) | <0.001 |

[a] Within each partner, the P values represent a difference <0.05 for the comparison between its training and test cohort.

[b] The P values represent a difference <0.05 for the comparison among three training cohorts from Partners 3, 4 and 6. The P values were adjusted using the Bonferroni correction.

[c] Data was reported based on values calculated at the latest hospital admission.

[d] Race and ethnicity were self-reported.

[e] Other races include American Indian or Alaska Native, Asian, Native Hawaiian or Pacific Islander, and multiracial.

[f] Data were reported based on postoperative complication status for each surgical procedure. When a patient had multiple surgeries during one admission, only the surgery with the highest intraoperative working units was used in the analysis.

*Comparison of Central Learning and Federated Learning Models*

We evaluated the performance of central learning and federated learning models both internally and externally using AUROC and AUPRC metrics (Table 2 and Table 3). Among the three federated learning algorithms tailored to neural network models—FedAvg, FedProx, and SCAFFOLD—SCAFFOLD consistently demonstrated comparable or superior performance in terms of AUROC and AUPRC across all outcomes and sites. When compared to the tree-based model, Federated XGBoost, SCAFFOLD achieved comparable or significantly better AUROC scores across almost all outcomes and sites, except for the AKI outcome in the external dataset (0.73 [95% CI, 0.71-0.77] vs 0.82 [95% CI, 0.79-0.84]). Similarly, SCAFFOLD achieved comparable or significantly higher AUPRC scores across most outcomes and sites, except for the AKI outcome for Partner 6 and the external dataset (Partner 6: 0.51 [95% CI, 0.49-0.52] vs 0.55 [95% CI, 0.54-0.57], External: 0.09 [95% CI, 0.06-0.12] vs 0.15 [95% CI, 0.12-0.19]). Thus, among all federated learning models, SCAFFOLD achieved the overall best performance. For simplicity, we only reported the SCAFFOLD performance in the following sections.

When compared with the central learning model, SCAFFOLD consistently achieved comparable or superior performance in terms of AUROC and AUPRC across all outcomes and sites, demonstrating its capability to provide strong predictive performance on non-independent and non-IID data distributions while preserving patient privacy. The performance of the SCAFFOLD model varied across different sites. For Partner 3, AUROC scores ranged from 0.82 (95% CI, 0.81-0.82) for AKI to 0.90 (95% CI, 0.90-0.91) for MV; AUPRC scores ranged from 0.18 (95% CI, 0.16-0.21) for in-hospital mortality to 0.69 (95% CI, 0.68-0.70) for ICU admission. For Partner 4, AUROC scores ranged from 0.94 (95% CI, 0.94-0.95) for ICU admission to 0.96 (95% CI, 0.96-0.96) for MV; AUPRC scores ranged from 0.03 (95% CI, 0.02-0.04) for in-hospital mortality to 0.43 (95% CI, 0.41-0.45) for ICU admission. For Partner 6, AUROC scores ranged from 0.79 (95% CI, 0.77-0.80) for ICU admission to 0.88 (95% CI, 0.86-0.89) for in-hospital mortality; AUPRC scores ranged from 0.12 (95% CI, 0.10-0.14) for hospital mortality to 0.51 (95% CI, 0.49-0.52) for AKI. For the external dataset, AUROC scores ranged from 0.73 (95% CI, 0.71-0.77) for AKI to 0.97 (95% CI, 0.96-0.99) for in-hospital mortality; AUPRC scores ranged from 0.01 (95% CI, 0.01-0.02) for ICU admission to 0.16 (95% CI, 0.08-0.29) for MV. Across all outcomes and sites, SCAFFOLD achieved generally high AUROC scores ranging from 0.73 to 0.97, with the highest score observed in MV and in-hospital mortality. However,

AUPRC scores, which account for class imbalance, varied more widely, highlighting differences in performance across sites and outcomes. This variability was particularly noticeable for extremely imbalanced outcomes, such as the in-hospital mortality outcome for Partner 4 and the ICU admission outcome for the external dataset.

**Table 2. Comparison of AUROC with 95% Confidence Interval for Central Learning and Federated Learning Models**

| Outcome | CL | FedAvg | FedProx | SCAFFOLD | Federated XGBoost |
|---|---|---|---|---|---|
| **Performance on Partner 3** | | | | | |
| ICU admission | **0.87 (0.87-0.88)** | 0.83 (0.82-0.83) | 0.81 (0.81-0.81) | 0.85 (0.85-0.86) | 0.78 (0.78-0.78) |
| MV | 0.89 (0.88-0.89) | 0.87 (0.87-0.88) | 0.87 (0.86-0.87) | **0.90 (0.90-0.91)** | 0.83 (0.83-0.84) |
| AKI | 0.78 (0.77-0.78) | 0.78 (0.77-0.79) | 0.77 (0.77-0.78) | **0.82 (0.81-0.82)** | 0.79 (0.79-0.80) |
| In-hospital mortality | 0.88 (0.87-0.89) | 0.88 (0.87-0.89) | 0.87 (0.86-0.88) | **0.89 (0.88-0.90)** | 0.87 (0.85-0.88) |
| **Performance on Partner 4** | | | | | |
| ICU admission | **0.94 (0.94-0.95)** | **0.94 (0.94-0.94)** | **0.94 (0.94-0.94)** | **0.94 (0.94-0.95)** | 0.92 (0.91-0.92) |
| MV | **0.96 (0.96-0.96)** | **0.96 (0.95-0.96)** | **0.96 (0.95-0.96)** | **0.96 (0.96-0.96)** | 0.94 (0.93-0.94) |
| AKI | 0.89 (0.88-0.89) | 0.93 (0.93-0.94) | 0.94 (0.93-0.94) | **0.95 (0.94-0.95)** | 0.90 (0.89-0.91) |
| In-hospital mortality | 0.89 (0.86-0.92) | **0.96 (0.95-0.97)** | **0.96 (0.95-0.97)** | **0.96 (0.94-0.97)** | 0.92 (0.88-0.95) |
| **Performance on Partner 6** | | | | | |
| ICU admission | 0.76 (0.75-0.77) | 0.77 (0.76-0.78) | 0.75 (0.74-0.76) | **0.79 (0.77-0.80)** | 0.75 (0.73-0.76) |
| MV | 0.78 (0.76-0.79) | 0.78 (0.77-0.79) | 0.76 (0.75-0.78) | **0.81 (0.80-0.82)** | 0.75 (0.74-0.77) |
| AKI | 0.76 (0.75-0.77) | 0.78 (0.77-0.79) | 0.77 (0.76-0.78) | **0.80 (0.79-0.81)** | **0.80 (0.79-0.81)** |
| In-hospital mortality | 0.83 (0.81-0.85) | 0.87 (0.86-0.89) | 0.86 (0.85-0.88) | **0.88 (0.86-0.89)** | **0.88 (0.86-0.89)** |
| **Performance on external validation dataset** | | | | | |
| ICU admission | 0.75 (0.65-0.83) | **0.77 (0.69-0.84)** | 0.75 (0.66-0.84) | **0.77 (0.69-0.84)** | 0.55 (0.45-0.64) |
| MV | **0.97 (0.95-0.98)** | 0.94 (0.92-0.96) | 0.83 (0.77-0.88) | 0.96 (0.95-0.98) | 0.86 (0.78-0.93) |
| AKI | 0.76 (0.73-0.79) | 0.76 (0.73-0.79) | 0.76 (0.73-0.78) | 0.73 (0.71-0.77) | **0.82 (0.79-0.84)** |
| In-hospital mortality | **0.98 (0.95-0.99)** | 0.96 (0.91-0.99) | 0.91 (0.85-0.97) | 0.97 (0.96-0.99) | 0.65 (0.39-0.88) |

Abbreviations. CL; central learning, ICU; intensive care unit, MV; mechanical ventilation, AKI; acute kidney injury. Highest performance for each row is bold.

**Table 3. Comparison of AUPRC with 95% Confidence Interval for Central Learning and Federated Learning Models**

| Outcome | CL | FedAvg | FedProx | SCAFFOLD | Federated XGBoost |
|---|---|---|---|---|---|
| **Performance on Partner 3** | | | | | |
| ICU admission | **0.71 (0.70-0.72)** | 0.64 (0.63-0.65) | 0.62 (0.62-0.63) | 0.69 (0.68-0.70) | 0.59 (0.58-0.60) |
| MV | 0.53 (0.52-0.54) | 0.50 (0.48-0.51) | 0.49 (0.47-0.50) | **0.56 (0.55-0.58)** | 0.47 (0.45-0.48) |
| AKI | 0.39 (0.37-0.40) | 0.39 (0.38-0.40) | 0.38 (0.37-0.40) | 0.46 (0.45-0.48) | **0.47 (0.46-0.49)** |
| In-hospital mortality | 0.18 (0.16-0.21) | 0.15 (0.13-0.17) | 0.14 (0.13-0.16) | 0.18 (0.16-0.21) | **0.20 (0.17-0.22)** |
| **Performance on Partner 4** | | | | | |
| ICU admission | **0.43 (0.41-0.45)** | 0.40 (0.38-0.42) | 0.42 (0.40-0.44) | **0.43 (0.41-0.45)** | 0.39 (0.37-0.41) |
| MV | **0.38 (0.35-0.40)** | 0.35 (0.33-0.38) | 0.36 (0.34-0.38) | **0.38 (0.36-0.40)** | 0.36 (0.33-0.39) |
| AKI | 0.15 (0.13-0.17) | 0.28 (0.25-0.31) | 0.29 (0.26-0.32) | **0.35 (0.32-0.38)** | 0.29 (0.26-0.32) |
| In-hospital mortality | 0.01 (0.01-0.03) | 0.02 (0.01-0.04) | **0.03 (0.02-0.07)** | **0.03 (0.02-0.04)** | 0.02 (0.01-0.05) |
| **Performance on Partner 6** | | | | | |
| ICU admission | 0.23 (0.22-0.25) | 0.23 (0.21-0.24) | 0.17 (0.16-0.19) | **0.25 (0.23-0.27)** | **0.25 (0.23-0.27)** |
| MV | 0.15 (0.13-0.17) | 0.15 (0.14-0.17) | 0.12 (0.11-0.13) | **0.19 (0.17-0.21)** | 0.15 (0.13-0.16) |
| AKI | 0.41 (0.40-0.43) | 0.47 (0.46-0.48) | 0.46 (0.44-0.47) | 0.51 (0.49-0.52) | **0.55 (0.54-0.57)** |
| In-hospital mortality | 0.08 (0.07-0.11) | 0.11 (0.09-0.13) | 0.09 (0.08-0.12) | 0.12 (0.10-0.14) | **0.14 (0.11-0.17)** |
| **Performance on external validation dataset** | | | | | |
| ICU admission | **0.01 (0.01-0.02)** | **0.01 (0.01-0.02)** | **0.01 (0.01-0.02)** | **0.01 (0.01-0.02)** | 0.00 (0.00-0.01) |
| MV | 0.13 (0.06-0.23) | 0.09 (0.04-0.18) | 0.08 (0.02-0.18) | **0.16 (0.08-0.29)** | 0.08 (0.04-0.18) |
| AKI | 0.08 (0.07-0.11) | 0.06 (0.05-0.08) | 0.06 (0.05-0.07) | 0.09 (0.06-0.12) | **0.15 (0.12-0.19)** |
| In-hospital mortality | 0.14 (0.02-0.45) | 0.04 (0.01-0.15) | 0.01 (0.00-0.02) | 0.12 (0.01-0.34) | **0.22 (0.00-0.49)** |

Abbreviations. CL; central learning, ICU; intensive care unit, MV; mechanical ventilation, AKI; acute kidney injury. Highest performance for each row is bold.

*Comparison of Local Learning and Federated Learning Models*

We internally and externally evaluated the performance of three local learning models (Partner 3, Partner 4 and Partner 6 models, developed using only their respective local data) and the federated learning model (SCAFFOLD) using AUROC and AUPRC metrics (Table 4 and Table 5). Local learning models tend to exhibit stronger performance on their respective local data but performed poorly at other sites. This is likely due to the test data at each site having a distribution more similar to the training data of the locally developed model than to the training data from other sites. Consequently, models developed using single-center data often have limited generalizability. However, compared to the best local learning model at each site, SCAFFOLD model demonstrated strong generalizability, with comparable or superior performance in terms of both AUROC and AUPRC. For the external evaluation, the Partner 3 model, developed using relatively outcome balanced data, achieved the best performance among all local learning models, and the SCAFFOLD model achieved comparable or superior performance.

**Table 4. Comparison of AUROC with 95% Confidence Interval for Local Learning and Federated Learning Models**

| Outcome | Model | Partner 3 Test Dataset | Partner 4 Test Dataset | Partner 6 Test Dataset | External Validation Dataset |
|---|---|---|---|---|---|
| ICU admission | Partner 3 model | **0.87 (0.86-0.87)** | 0.86 (0.85-0.86) | 0.74 (0.73-0.75) | **0.79 (0.71-0.85)** |
| | Partner 4 model | 0.64 (0.63-0.64) | **0.95 (0.94-0.95)** | 0.62 (0.61-0.64) | 0.71 (0.61-0.79) |
| | Partner 6 model | 0.77 (0.76-0.77) | 0.66 (0.65-0.67) | **0.80 (0.78-0.81)** | 0.74 (0.66-0.81) |
| | SCAFFOLD | 0.85 (0.85-0.86) | 0.94 (0.94-0.95) | 0.79 (0.77-0.80) | 0.77 (0.69-0.84) |
| MV | Partner 3 model | 0.89 (0.89-0.90) | 0.89 (0.88-0.90) | 0.76 (0.74-0.77) | 0.95 (0.93-0.96) |
| | Partner 4 model | 0.68 (0.67-0.69) | **0.96 (0.96-0.96)** | 0.65 (0.64-0.67) | 0.75 (0.65-0.85) |

| Outcome | Model | Partner 3 Test Dataset | Partner 4 Test Dataset | Partner 6 Test Dataset | External Validation Dataset |
|---|---|---|---|---|---|
| | Partner 6 model | 0.79 (0.79-0.80) | 0.69 (0.68-0.70) | 0.80 (0.78-0.81) | 0.90 (0.87-0.93) |
| | SCAFFOLD | **0.90 (0.90-0.91)** | **0.96 (0.96-0.96)** | **0.81 (0.80-0.82)** | **0.96 (0.95-0.98)** |
| AKI | Partner 3 model | 0.81 (0.80-0.81) | 0.75 (0.73-0.76) | 0.73 (0.73-0.74) | **0.78 (0.76-0.81)** |
| | Partner 4 model | 0.62 (0.61-0.63) | **0.95 (0.95-0.96)** | 0.62 (0.61-0.63) | 0.66 (0.62-0.69) |
| | Partner 6 model | 0.74 (0.74-0.75) | 0.69 (0.68-0.71) | 0.79 (0.79-0.80) | 0.73 (0.70-0.77) |
| | SCAFFOLD | **0.82 (0.81-0.82)** | **0.95 (0.94-0.95)** | **0.80 (0.79-0.81)** | 0.73 (0.71-0.77) |
| In-hospital mortality | Partner 3 model | **0.91 (0.90-0.92)** | 0.64 (0.56-0.72) | 0.80 (0.78-0.82) | **0.97 (0.95-0.99)** |
| | Partner 4 model | 0.71 (0.69-0.73) | 0.95 (0.94-0.97) | 0.66 (0.63-0.68) | 0.68 (0.48-0.85) |
| | Partner 6 model | 0.81 (0.80-0.83) | 0.73 (0.66-0.80) | 0.87 (0.86-0.89) | 0.91 (0.80-0.97) |
| | SCAFFOLD | 0.89 (0.88-0.90) | **0.96 (0.94-0.97)** | **0.88 (0.86-0.89)** | **0.97 (0.96-0.99)** |

Abbreviations. CL; central learning, ICU; intensive care unit, MV; mechanical ventilation, AKI; acute kidney injury. Highest performance for each outcome and each column is bold.

**Table 5. Comparison of AUPRC with 95% Confidence Interval for Local Learning and Federated Learning Models**

| Outcome | Model | Partner 3 Test Dataset | Partner 4 Test Dataset | Partner 6 Test Dataset | External Validation Dataset |
|---|---|---|---|---|---|
| ICU admission | Partner 3 model | 0.70 (0.69-0.71) | 0.26 (0.24-0.28) | 0.22 (0.20-0.23) | **0.01 (0.01-0.03)** |
| | Partner 4 model | 0.42 (0.41-0.43) | **0.46 (0.44-0.48)** | 0.10 (0.09-0.10) | **0.01 (0.00-0.01)** |
| | Partner 6 model | 0.55 (0.54-0.56) | 0.10 (0.09-0.11) | **0.28 (0.26-0.30)** | **0.01 (0.00-0.01)** |
| | SCAFFOLD | 0.69 (0.68-0.70) | 0.43 (0.41-0.45) | 0.25 (0.23-0.27) | **0.01 (0.01-0.02)** |
| MV | Partner 3 model | 0.55 (0.53-0.56) | 0.20 (0.18-0.22) | 0.14 (0.13-0.16) | 0.07 (0.03-0.12) |
| | Partner 4 model | 0.25 (0.24-0.26) | **0.39 (0.37-0.42)** | 0.07 (0.06-0.08) | 0.10 (0.03-0.23) |
| | Partner 6 model | 0.36 (0.35-0.38) | 0.07 (0.06-0.08) | 0.17 (0.15-0.19) | 0.02 (0.01-0.03) |

| Outcome | Model | Partner 3 Test Dataset | Partner 4 Test Dataset | Partner 6 Test Dataset | External Validation Dataset |
|---|---|---|---|---|---|
| | SCAFFOLD | **0.56 (0.55-0.58)** | 0.38 (0.36-0.40) | **0.19 (0.17-0.21)** | **0.16 (0.08-0.29)** |
| AKI | Partner 3 model | 0.45 (0.43-0.46) | 0.10 (0.09-0.12) | 0.38 (0.36-0.39) | 0.07 (0.06-0.09) |
| | Partner 4 model | 0.22 (0.22-0.23) | 0.32 (0.29-0.35) | 0.28 (0.27-0.29) | 0.05 (0.04-0.06) |
| | Partner 6 model | 0.33 (0.32-0.35) | 0.09 (0.08-0.11) | 0.48 (0.47-0.50) | 0.06 (0.05-0.08) |
| | SCAFFOLD | **0.46 (0.45-0.48)** | **0.35 (0.32-0.38)** | **0.51 (0.49-0.52)** | **0.09 (0.06-0.12)** |
| In-hospital mortality | Partner 3 model | **0.22 (0.20-0.25)** | 0.01 (0.00-0.03) | 0.08 (0.06-0.10) | **0.15 (0.01-0.38)** |
| | Partner 4 model | 0.05 (0.05-0.06) | **0.03 (0.02-0.09)** | 0.03 (0.02-0.03) | 0.00 (0.00-0.01) |
| | Partner 6 model | 0.09 (0.08-0.10) | 0.01 (0.00-0.01) | 0.10 (0.09-0.12) | 0.01 (0.00-0.02) |
| | SCAFFOLD | 0.18 (0.16-0.21) | **0.03 (0.02-0.04)** | **0.12 (0.10-0.14)** | 0.12 (0.01-0.34) |

Abbreviations. CL; central learning, ICU; intensive care unit, MV; mechanical ventilation, AKI; acute kidney injury. Highest performance for each outcome and each column is bold.

*Sensitivity Analysis*

At each partner, we fine-tuned the federated learning model using its local feature, the surgeon's identity, and evaluated its performance (Table 6). Across almost all sites and all outcomes, the fine-tuned federated learning model demonstrated consistent comparable and slightly better AUROC and AUPRC scores. Specifically, for Partner 3, the fine-tuned model showed improvements over the federated learning model in outcomes such as ICU admission (AUROC 0.87 [95% CI, 0.86-0.87] vs. 0.85 [95% CI, 0.85-0.86]; AUPRC 0.71 [95% CI, 0.70-0.72] vs. 0.69 [95% CI, 0.68-0.70]), AKI (AUPRC 0.47 [95% CI, 0.46-0.49] vs. 0.46 [95% CI, 0.45-0.48]), and in-hospital mortality (AUROC 0.90 [95% CI, 0.89-0.90] vs. 0.89 [95% CI, 0.88-0.90]; AUPRC 0.20 [95% CI, 0.18-0.23] vs. 0.18 [95% CI, 0.16-0.21]). The improvements were also reflected in slightly higher specificity and PPV values. For Partner 4, the finetuned model showed an increase in AUPRC scores for ICU admission (0.46 [95% CI, 0.44-0.48] vs. 0.43

[95% CI, 0.41-0.45]) and MV (0.41 [95% CI, 0.39-0.43] vs. 0.38 [95% CI, 0.36-0.40]), while other metrics remained similar. For Partner 6, the fine-tuned model provided improvements for ICU admission (AUROC 0.80 [95% CI, 0.79-0.81] vs. 0.79 [95% CI, 0.77-0.80]; AUPRC 0.28 [95% CI, 0.26-0.30] vs. 0.25 [95% CI, 0.23-0.27]), MV (AUPRC 0.20 [95% CI, 0.18-0.22] vs. 0.19 [95% CI, 0.17-0.21]), AKI (AUROC 0.81 [95% CI, 0.80-0.81] vs. 0.80 [95% CI, 0.79-0.81]; AUPRC 0.53 [95% CI, 0.51-0.54] vs. 0.51 [95% CI, 0.49-0.52]), and in-hospital mortality (AUPRC 0.13 [95% CI, 0.11-0.16] vs. 0.12 [95% CI, 0.10-0.14]). The model also showed slight improvements in specificity, sensitivity and PPV metrics for most outcomes.

**Table 6. Sensitivity Analysis Fine-tuning the Federated Learning SCAFFOLD Model by Adding Personalized Feature (the Surgeon's Identity): Model Performance Measurements with 95% Confidence Interval**

| Outcome | Model | AUROC | AUPRC | Sensitivity | Specificity | PPV | NPV |
|---|---|---|---|---|---|---|---|
| **Performance on Partner 3** | | | | | | | |
| ICU admission | FL | 0.85 (0.85-0.86) | 0.69 (0.68-0.70) | 0.78 (0.75-0.80) | 0.77 (0.74-0.79) | 0.53 (0.51-0.55) | 0.91 (0.90-0.92) |
| | FT | **0.87 (0.86-0.87)** | **0.71 (0.70-0.72)** | 0.78 (0.76-0.81) | **0.81 (0.77-0.82)** | **0.57 (0.54-0.59)** | 0.91 (0.91-0.92) |
| MV | FL | 0.90 (0.90-0.91) | 0.56 (0.55-0.58) | 0.86 (0.83-0.87) | 0.81 (0.79-0.84) | 0.37 (0.35-0.40) | 0.98 (0.97-0.98) |
| | FT | 0.90 (0.90-0.91) | 0.56 (0.55-0.58) | 0.86 (0.84-0.88) | 0.81 (0.79-0.82) | 0.37 (0.35-0.39) | 0.98 (0.98-0.98) |
| AKI | FL | 0.82 (0.81-0.82) | 0.46 (0.45-0.48) | **0.78 (0.70-0.81)** | 0.70 (0.68-0.78) | 0.30 (0.29-0.34) | 0.95 (0.94-0.96) |
| | FT | 0.82 (0.81-0.82) | **0.47 (0.46-0.49)** | 0.74 (0.72-0.77) | **0.74 (0.71-0.76)** | **0.32 (0.30-0.33)** | 0.95 (0.94-0.95) |
| In-hospital mortality | FL | 0.89 (0.88-0.90) | 0.18 (0.16-0.21) | **0.84 (0.78-0.88)** | 0.79 (0.74-0.84) | 0.08 (0.07-0.10) | 1.00 (0.99-1.00) |
| | FT | **0.90 (0.89-0.90)** | **0.20 (0.18-0.23)** | 0.81 (0.79-0.88) | **0.82 (0.75-0.83)** | **0.09 (0.07-0.10)** | 1.00 (0.99-1.00) |
| **Performance on Partner 4** | | | | | | | |

| Outcome | Model | AUROC | AUPRC | Sensitivity | Specificity | PPV | NPV |
|---|---|---|---|---|---|---|---|
| ICU admission | FL | 0.94 (0.94-0.95) | 0.43 (0.41-0.45) | 0.89 (0.84-0.91) | 0.86 (0.84-0.90) | 0.20 (0.18-0.26) | 0.99 (0.99-1.00) |
| | FT | **0.95 (0.94-0.95)** | **0.46 (0.44-0.48)** | 0.89 (0.86-0.93) | 0.86 (0.82-0.88) | 0.20 (0.18-0.23) | 0.99 (0.99-1.00) |
| MV | FL | 0.96 (0.96-0.96) | 0.38 (0.36-0.40) | **0.91 (0.89-0.92)** | 0.92 (0.92-0.94) | 0.22 (0.21-0.26) | 1.00 (1.00-1.00) |
| | FT | 0.96 (0.96-0.97) | **0.41 (0.39-0.43)** | 0.90 (0.89-0.93) | **0.93 (0.91-0.93)** | **0.23 (0.19-0.25)** | 1.00 (1.00-1.00) |
| AKI | FL | 0.95 (0.94-0.95) | **0.35 (0.32-0.38)** | **0.91 (0.87-0.94)** | 0.85 (0.83-0.89) | 0.10 (0.09-0.13) | **1.00 (1.00-1.00)** |
| | FT | 0.95 (0.94-0.95) | 0.33 (0.30-0.36) | 0.90 (0.87-0.94) | **0.86 (0.82-0.88)** | **0.11 (0.09-0.13)** | 1.00 (1.00-1.00) |
| In-hospital mortality | FL | 0.96 (0.94-0.97) | 0.03 (0.02-0.04) | 0.94 (0.85-1.00) | **0.85 (0.76-0.95)** | 0.01 (0.00-0.02) | 1.00 (1.00-1.00) |
| | FT | 0.96 (0.95-0.97) | 0.03 (0.02-0.05) | **0.97 (0.89-1.00)** | 0.84 (0.80-0.94) | 0.01 (0.00-0.01) | 1.00 (1.00-1.00) |
| **Performance on Partner 6** | | | | | | | |
| ICU admission | FL | 0.79 (0.77-0.80) | 0.25 (0.23-0.27) | 0.69 (0.65-0.81) | 0.74 (0.62-0.77) | 0.17 (0.14-0.19) | 0.97 (0.96-0.98) |
| | FT | **0.80 (0.79-0.81)** | **0.28 (0.26-0.30)** | **0.70 (0.65-0.77)** | **0.75 (0.68-0.79)** | **0.18 (0.16-0.20)** | 0.97 (0.97-0.97) |
| MV | FL | 0.81 (0.80-0.82) | 0.19 (0.17-0.21) | **0.75 (0.70-0.81)** | 0.73 (0.68-0.79) | 0.12 (0.10-0.13) | 0.98 (0.98-0.99) |
| | FT | 0.81 (0.80-0.82) | **0.20 (0.18-0.22)** | 0.72 (0.70-0.82) | **0.77 (0.67-0.79)** | **0.13 (0.10-0.14)** | 0.98 (0.98-0.99) |
| AKI | FL | 0.80 (0.79-0.81) | 0.51 (0.49-0.52) | 0.73 (0.69-0.75) | 0.71 (0.70-0.75) | 0.36 (0.35-0.38) | 0.92 (0.92-0.93) |
| | FT | **0.81 (0.80-0.81)** | **0.53 (0.51-0.54)** | 0.73 (0.67-0.77) | **0.73 (0.69-0.78)** | **0.37 (0.35-0.41)** | 0.92 (0.92-0.93) |
| In-hospital mortality | FL | 0.88 | 0.12 | 0.82 (0.77-0.92) | **0.78 (0.69-0.84)** | 0.05 | 1.00 |

| Outcome | Model | AUROC | AUPRC | Sensitivity | Specificity | PPV | NPV |
|---|---|---|---|---|---|---|---|
| | | (0.86-0.89) | (0.10-0.14) | | | (0.04-0.07) | (1.00-1.00) |
| | FT | 0.88 (0.87-0.90) | **0.13 (0.11-0.16)** | **0.84 (0.76-0.92)** | 0.76 (0.68-0.83) | 0.05 (0.04-0.07) | 1.00 (1.00-1.00) |

Abbreviations. FL; federated learning, FT; fine-tuning, ICU; intensive care unit, MV; mechanical ventilation, AKI; acute kidney injury. Highest performance for each outcome is bold.

## Discussion

In this study, we developed and internally and externally validated federated learning models to predict the risk of major postoperative complications and mortality using a large multicenter OneFlorida Data Trust dataset. The federated learning SCAFFOLD model demonstrated the best performance for our non-IID data and proved to be more robust and generalizable than local models. The federated learning model also showed comparable and slightly superior performance to the central learning model while maintaining data privacy.

Our study highlights several key advantages of federated learning over traditional central learning and local learning approaches. First, the development of federated learning models allows multiple centers to participate, leveraging large and diverse patient populations. This mitigates the potential biases introduced by smaller, single-center datasets and ensures that the model can generalize well across different populations and healthcare settings, which is crucial for real-world clinical applications. This capability is demonstrated in our study and supported by many other studies[11,16,24-26]. For example, Dayan et al.[24] proposed a federated learning model to predict the future oxygen requirements for COVID-19 patients using EHR data from 20 institutions. The model provided a 38% increase in generalizability, as measured by AUROC scores, compared to local models. Similarly, Vaid et al.[11] developed federated learning models to predict mortality in COVID-19 patients within 7 days using data from 5 hospitals, which outperformed all local models. Second, the federated learning approach aligns with current ethical and legal standards for data privacy, addressing a major barrier in the sharing and utilization of medical data. Furthermore, the use of the federated learning SCAFFOLD model specifically addresses challenges associated with non-IID data distributions, which are common in multicenter studies.

Our findings highlight the feasibility of implementing federated learning in a practical healthcare setting. The model's robust predictive performance, while preserving data privacy, demonstrates its potential for integration into clinical decision support systems. By utilizing only preoperative data, our model enables early assessment of surgical risk, helping clinicians identify high-risk patients and tailor perioperative care to reduce the likelihood of complications. For example, patients at high risk of postoperative AKI or cardiovascular events could benefit from preoperative interventions such as optimizing fluid status, controlling blood pressure, and managing anemia.[27] Additionally, for patients with diabetes, controlling weight and adjusting lifestyle can significantly enhance their overall surgical outcomes.[28] Early identification of high-risk patients also facilitates improved perioperative planning and resource allocation. It allows for the optimization of operating room and ICU bed schedules, prioritization of resources for the most vulnerable patients, and the preparation of necessary equipment and medications. By implementing these measures, healthcare systems can enhance patient outcomes and reduce the overall burden on healthcare infrastructure.

Sensitivity analysis highlights the potential of combining federated learning and local learning approaches. While using routinely collected features for model development can increase the generalizability of models, adding additional features may provide further valuable insights. For example, features such as surgeons' previous performances in relation to his case-mix, and patients' social determinants of health (e.g., poverty and inequality), have been increasingly recognized for their impact on health.[29,30] Adding these features to model development allows the model to provide more personalized predictions and account for local variations that might not be captured by the federated learning model alone. Additionally, if the performance of the federated model drops due to data drift in some centers, fine-tuning the federated learning model in those centers with updated data allows the quick adaptation.

The study has several limitations. First, the dataset used lacks certain granular details of surgeries (such as the exact start and end times) and the types of anesthesia administered, which could introduce potential bias into our results. Additionally, the absence of data on factors such as patient locations (stations) and respiratory device usage limits the scope of our outcomes. Second, while our study simulated the federated learning environment, it did not fully replicate the practical challenges of real-world implementation. For example, variations in EHR data cleaning, labeling, and standardization practices across different centers were not addressed. Additionally, technical challenges such as network latency, data synchronization, and differences in computational power among participating nodes were not evaluated. These factors could significantly influence the feasibility and performance of federated learning in an actual deployment scenario. Future work should emphasize the creation of a multicenter surgical dataset with clear provenance, employing common data models and including comprehensive elements specific to surgeries from diverse, multi-institutional sources.

## Conclusions

We developed robust, generalizable, and privacy-preserving predictive models for major postoperative complications and mortality using a large multicenter dataset. Further implementation studies are needed to validate the federated learning platform across different healthcare systems and to assess the clinical impact of these models on patient care.